\documentclass{article}
\usepackage{amsmath,graphicx,mlspconf}
\usepackage{xcolor}
\usepackage{comment}
\usepackage{amssymb}
\usepackage{cite}
\usepackage{multirow}
\usepackage{hyperref} 
\usepackage{orcidlink}
\copyrightnotice{979-8-3503-2411-2/25/\$31.00 {\copyright}2025 IEEE}

\toappear{2026 IEEE International Workshop on Machine Learning for Signal Processing, Sep.\ 28-- Oct.\ 1, 2026, Atlanta, USA}

\title{A 2-Block Architecture for Real-Time EEG Gait Decoding: A Pilot Study}
\name{Anonymous\thanks{Anonymous.}}
\address{Anonymous}

\name{Shantanu Sarkar$^{1*}$\orcidlink{0000-0002-5125-6925} \qquad
      Saurabh Prasad$^{1}$ \orcidlink{0000-0003-3729-9360} \qquad
      Jose Luis Contreras-Vidal$^{1}$ \orcidlink{0000-0002-6499-1208}
      \thanks{This work was supported by NSF IUCRC BRAIN Center (\#2137255).}}
\address{$^{1}$ IUCRC BRAIN Center, Cullen College of Engineering, University of Houston, Houston, TX, USA \\
         $^{*}$ \href{mailto:shantanu75@gmail.com}{shantanu75@gmail.com}}

\begin{document}

\maketitle

\begin{abstract}
Closed-loop lower-limb exoskeleton control via Electroencephalography (EEG) remains limited by motion artifacts, low signal-to-noise ratio, and binary gait formulations that fail to capture full cortical gait complexity. We propose a 2-block Brain-Computer Interface (BCI) architecture: a trainable session-specific Feature Extraction Block with real-time artifact suppression and 
multi-domain feature extraction, coupled with a Decoder Block built on a novel Polynomial Time-Varying Layer (PolyTVL)+LSTM for four-state gait classification (Stand, Initiate, Execute, Terminate). Ablation confirmed $v01$ (PolyTVL+LSTM) outperformed all variants (validation MCC: 0.435, gap: 0.187), with consistent EEG feature discriminability across ROIs and sub-bands ($p<0.05$). Closed-loop deployment with $v01$ achieved 55.3\% (Rex-assisted) and 52.7\% (volitional) gait initiation success, with mean end-to-end processing time of 70.5~ms ($\pm$41.5), validating real-time feasibility in this pilot study.
\end{abstract}

\newcommand{\cem}[1]{\textcolor{blue}{cem: #1}}
\section{Introduction}\label{sec:intro}
Electroencephalography (EEG)-controlled powered exoskeleton locomotion is a clinically relevant branch of Brain-Computer Interface (BCI) research, with the primary goal of restoring motor and ambulatory function in individuals with paresis through repetitive, task-specific training~\cite{Contreras2016}. Since the early demonstration of lower-limb movement-related mu rhythm modulation in EEG~\cite{Pfurtscheller1994}, a growing body of work has established the feasibility of lower-limb Motor Imagery (MI) and movement intention decoding~\cite{Ortiz2023,MorenoCastelblanco2025}. However, the majority of these studies are conducted offline, in which neural data are recorded and analyzed post hoc. Closed-loop (online) BCI refers to real-time decoding of neural activity to drive an external device, with feedback that continuously shapes ongoing brain activity~\cite{Ortiz2023}. Despite its importance, only a handful of lower-limb BCI studies have demonstrated true closed-loop (CL) operation, underscoring a critical gap between offline feasibility and deployable real-time systems.
\\
A key challenge in EEG decoding is distinguishing genuine neural biomarkers of motor intent from movement-related artifacts~\cite{McDermott2022}, which overlap with neural signals during gait~\cite{Ortiz2023} -- making offline validation an insufficient proxy. Furthermore, most lower-limb BCI studies treat the problem as binary (walk/stop)~\cite{Ortiz2023,MorenoCastelblanco2025}, failing to capture the full cortical complexity of gait — a behavior requiring coordinated sensory-motor interactions~\cite{Roeder2024}. Volitional exoskeleton control therefore demands real-time artifact suppression and a multi-state formulation spanning at least four gait states: \textit{Stand}, \textit{Initiate}, \textit{Execute}, and \textit{Terminate}.
\\
The temporal nature of the gait cycle makes long-range sequence modeling essential. He et al. demonstrated successful decoding of lower-limb kinematics from EEG via an Unscented Kalman Filter (UKF) during NASA X1 exoskeleton-assisted walking in a stroke survivor~\cite{He2014}. Nakagome et al. found that the UKF achieved better early convergence at smaller tap sizes, whereas recurrent architectures outperformed other methods at larger tap sizes~\cite{Nakagome2020}. Tortora et al. also showed that a two-layer Long Short-Term Memory (LSTM) achieves AUC $>90$\% for swing and stance classification from mobile EEG~\cite{Tortora2020}.
\\
Recent developments in structured state-space models (SSMs) — S4, DSS, and S4D — enable efficient parallel sequence modeling via convolutional SSM kernels~\cite{Gu2022}. Guo et al. reported 80.59\% and 84.42\% MI decoding accuracy on BCI Competition IV 2a and 2b data using the Mamba-based SSM~\cite{Guo2025}; however, SSMs remain linear time-invariant (LTI) systems, constrained in modeling the nonlinear and non-stationary dynamics of real EEG. To overcome this, we propose the Polynomial Time-Varying Layer (PolyTVL), coupled with average pooling and an LSTM-based sequence summarizer to capture position-specific nonlinear dynamics.
\\
A known challenge in EEG-based BCI is the low signal-to-noise ratio from physiological and non-physiological artifacts~\cite{Rashmi2022}. Conventional preprocessing pipelines — Independent Component Analysis (ICA), which imposes high computational overhead, and ASR, which relies on heuristic thresholding — preclude real-time CL BCI application~\cite{nASR2026}. Therefore, to achieve real-time denoising, we incorporated adaptive ocular artifact removal via $H^\infty$~\cite{HInf2016}, band-pass filtering (BPF: 0.5 -- 30 Hz), and a neural Artifact Subspace Reconstruction layer (nASR)~\cite{nASR2026}.
\\
EEG-based MI decoding relies on discriminative features representing combinations of three primary domains: time, frequency, and spatial~\cite{Singh2021}. Zhang et al. showed the importance of weighted Region of Interest (ROI)-based spatial domain for lower-limb motor control~\cite{Zhang2017}. Building on this, we introduce ROI-based depth-wise convolution across nine anatomically defined ROIs spanning 28 channels. 
Time-frequency domain features extracted via Wavelet Transform are common practice in EEG-based decoding~\cite{MorenoCastelblanco2025,Pooja2022}; accordingly, we employ an adaptive Redundant Discrete Wavelet Transform (RDWT) filtering layer, implemented as a trainable neural network layer within the computational graph~\cite{SarkarRDWT2026} — to decompose each signal (sampled at 100\,Hz) into four sub-bands: $\delta,\theta$ (0--6.25\,Hz), $\alpha$ (6.25--12.5\,Hz), $\beta$ (12.5--25\,Hz), and $\gamma$ ($>$25\,Hz). The $\gamma$ sub-band is discarded prior to decoding as EMG contamination is typically prominent above 25 Hz~\cite{Pope2022}, leaving three physiologically relevant spectral representations. Together, these stages constitute the `Feature Extraction Block', delivering clean multi-domain spatial-spectral-temporal features for downstream decoding.
\\
The `Decoder Block' receives the output of `Feature Extraction Block' through three parallel branches (one per frequency band) — each comprising a PolyTVL, average pooling, and an LSTM-based sequence summarizer. The three band-specific representations are concatenated and passed through a two-stage dense layer for four-class gait state decoding, directly mapping to discrete volitional commands for safe exoskeleton control.
\\
The proposed framework addresses three core requirements for a deployable CL lower-limb BCI:
\\
(1)\ Real-time artifact suppression via $H^\infty$, BPF, nASR, and trainable RDWT filtering.
\\
(2)\ Multi-domain feature extraction through ROI-based depth-wise spatial filtering and adaptive RDWT coefficient filtering, yielding time-frequency-spatial domain representations.
\\
(3)\ Temporal sequence modeling of time-varying nonlinear dynamics via PolyTVL, coupled with average pooling and an LSTM-based sequence summarizer, enabling robust four-state gait classification for CL lower-limb exoskeleton control.
\vspace{-10pt}
\section{Materials and Methods}
\vspace{-5pt}
\subsection{Experiment Paradigm}
The experimental paradigm approved by the University of Houston IRB (STUDY00003848) consisted of 10 Sessions (Ses.) divided into two phases: an open-loop (OL) phase (Ses.~1--5) and a mixed open- and CL phase (Ses.~6--10).
\\
\textbf{Ses.~1–5 (OL Phase):} Each Ses. comprised five runs of 20 gait cycles, each preceded by a rest of 2-sec. A 200 ms audio cue at 12-sec intervals triggered each cycle, followed by exoskeleton step initiation after a random delay of 0.5--1.5 sec ($\approx9$ sec per step). Participants performed MI of a complete step, starting with the right leg, in synchronization with the audio cue and exoskeleton. Four gait states were defined as: \textit{Initiate} (2-sec post-cue), \textit{Execute} (motion period), \textit{Terminate} (2-sec pre-halt), and \textit{Stand} (rest). The 2-sec duration is consistent with the Bereitschaftspotential (BP) onset $\sim$2s before movement~\cite{shibasaki2006}. Steps 1--10 and 16--20 were used for training, and steps 11--15 for validation (75:25 split), ensuring non-overlapping windows to prevent data leakage. The final decoder was then trained on features aggregated across all five session-specific `Feature Extraction Blocks'.
\\
\textbf{Ses.~6-10 (Mixed OL / CL Phase):} Each session comprised nine runs (2-sec rest each): OL Runs~1-3(20 steps) and CL Runs~4-9 (10 steps), totaling 2 min to comply with the 2-min walk test~\cite{Bohannon2015}.
\\
\textit{Runs~1-3 (OL):} Identical to Ses.~1–5; data retrained the `Feature Extraction Block' (75:25 split) while the `Decoder Block' remained frozen.
\\
\textit{Runs 4-6 (CL-Rex):} Ten gait cycles at 12-sec intervals, with exoskeleton steps triggered in real time by the session-specific `Feature Extraction Block' stacked with the frozen `Decoder Block'.
\\
\textit{Runs~7-9 (CL-Manual):} Identical to Runs~4–6 but with manual stepping without exoskeleton assistance, establishing ground-truth (volitional intent).
\\
For CL Runs, a valid initiation is considered if $\geq$2 \textit{Initiate} predictions are detected within 10 decoding windows (2-sec, 200ms stride); a null decoder yields only 1.67 expected hits (16.7\% chance rate), below the two-prediction threshold.
\begin{figure}[!ht]
    \centering
    \includegraphics[width=\linewidth]{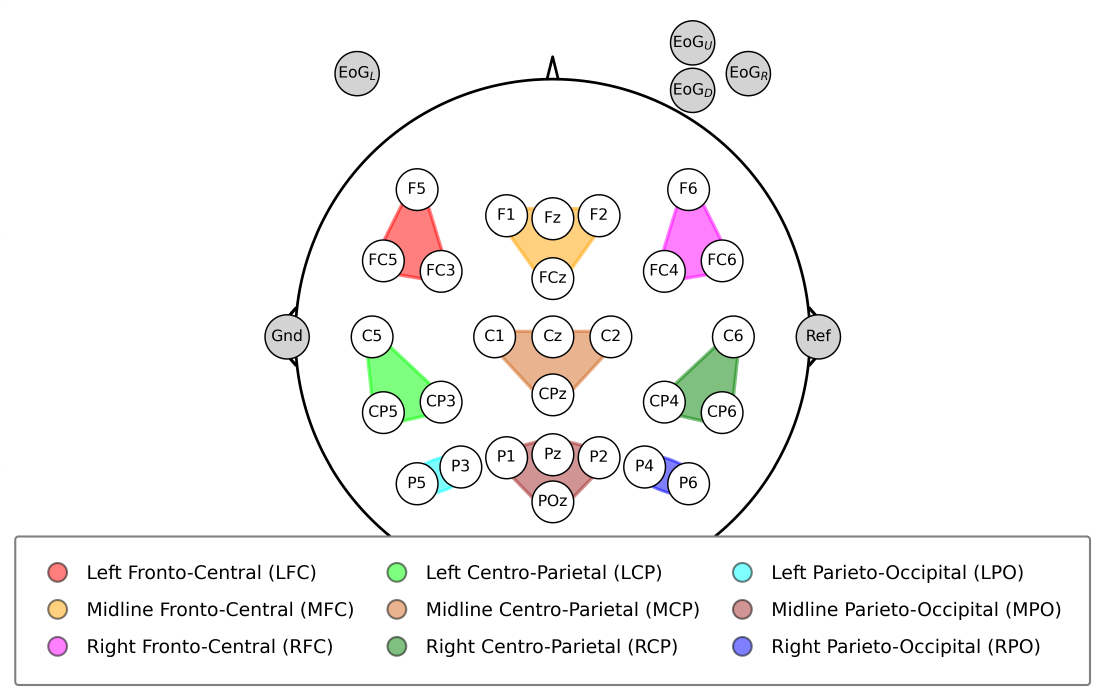}
    \vspace{-10pt}
    \caption{Electrode layout: 28 EEG channels across 9 ROIs}
    \label{Fig1}
    \vspace{-10pt}
\end{figure}

\vspace{-3pt}
\subsection{Participant and Data Acquisition}
For the pilot study, a healthy right-handed male participant (S1: 22\,Y) was recruited. During each Ses., EEG from 9 ROIs (28 channels) and 4 electrooculography (EOG) channels were recorded using the Brain Products actiCAP and MOVE system (Brain Products GmbH, Germany) at 100\,Hz, with ground and reference electrodes placed on the left and right earlobes, respectively. Electrode positions are shown in Fig.~\ref{Fig1}. The study used the Rex lower-limb exoskeleton (Rex Bionics Ltd., Auckland, New Zealand), which provided full locomotion assistance via 10 actuators. Additionally, electromyography (EMG) and inertial measurement unit (IMU) data were acquired using three Trigno Avanti sensors (Delsys Inc., Natick, MA, USA) placed on the left and right tibialis anterior (TA) muscles and the back of the neck for future analysis. Data acquisition was synchronized via a custom trigger box, while the Python-based control and acquisition system issued trigger markers and controlled the Rex in real time.
\vspace{-3pt}
\subsection{Preprocessing}\label{sec:Preprocessing}
EEG samples were first processed using $H^{\infty}$, leveraging EOG references to suppress ocular artifacts\cite{HInf2016}. The sliding window of 256 samples (2.56~sec) with a 20-sample stride (200~ms) was then applied. Each window was filtered using a zero-phase $2^{nd}$-Order Butterworth BPF (0.5--30~Hz), followed by a zero-phase $4^{th}$-Order Chebyshev Type~II low-pass filter (30~Hz) to attenuate high-frequency noise.
\\
Following filtering, each window was normalized to channel-wise zero mean. During training, windows in which all channel amplitudes remained within $\pm 3.5~\sigma$ were considered artifact-free and used to estimate session-specific per-channel reference $\mu$ and $\sigma$, which were then applied for channel-wise z-score normalization prior to passing to the nASR layer \cite{nASR2026}.

\begin{figure*}[!ht]
    \centering
    \includegraphics[width=0.9\textwidth]{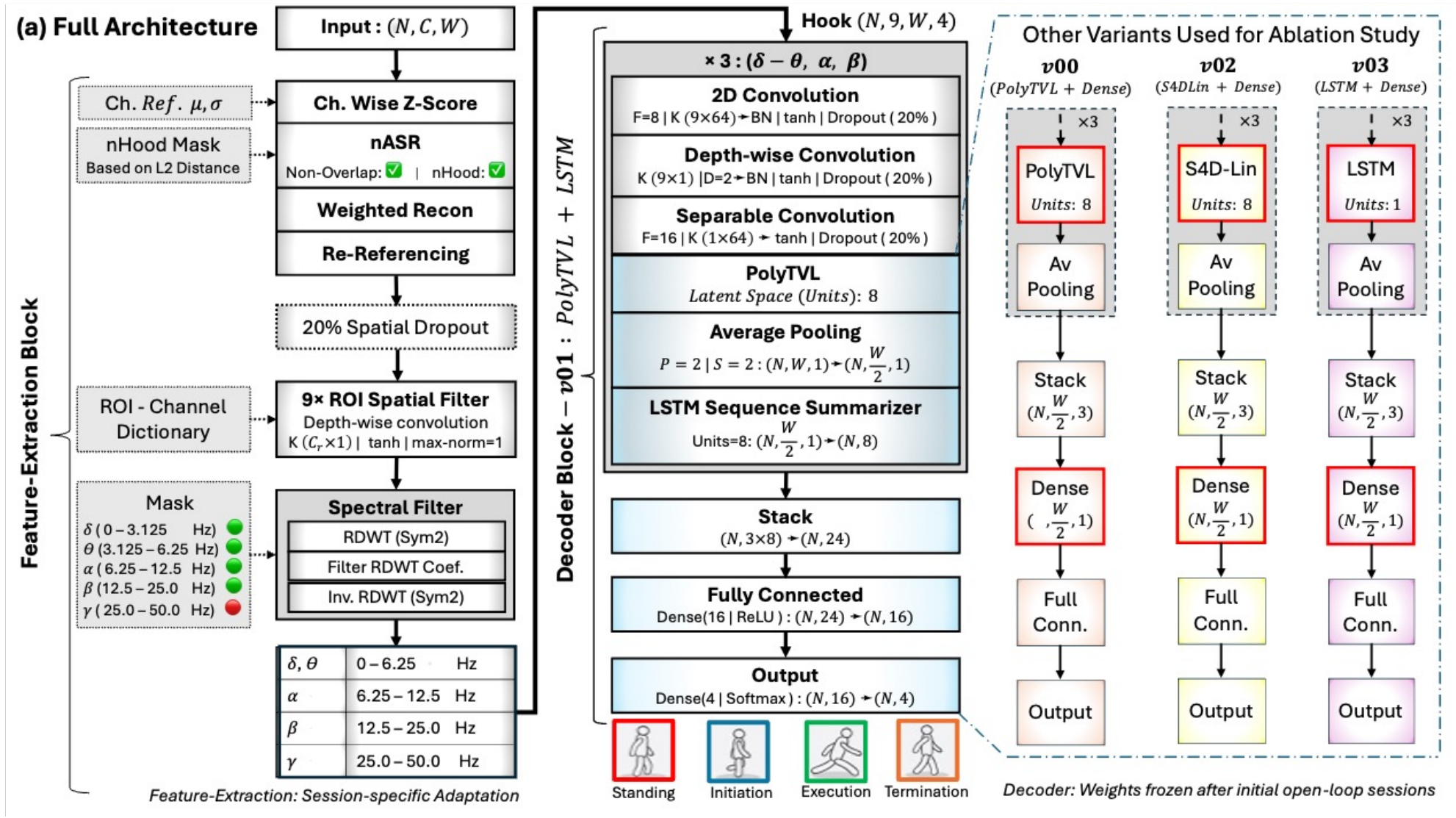}
    \vspace{2pt}
    \includegraphics[width=0.7\textwidth]{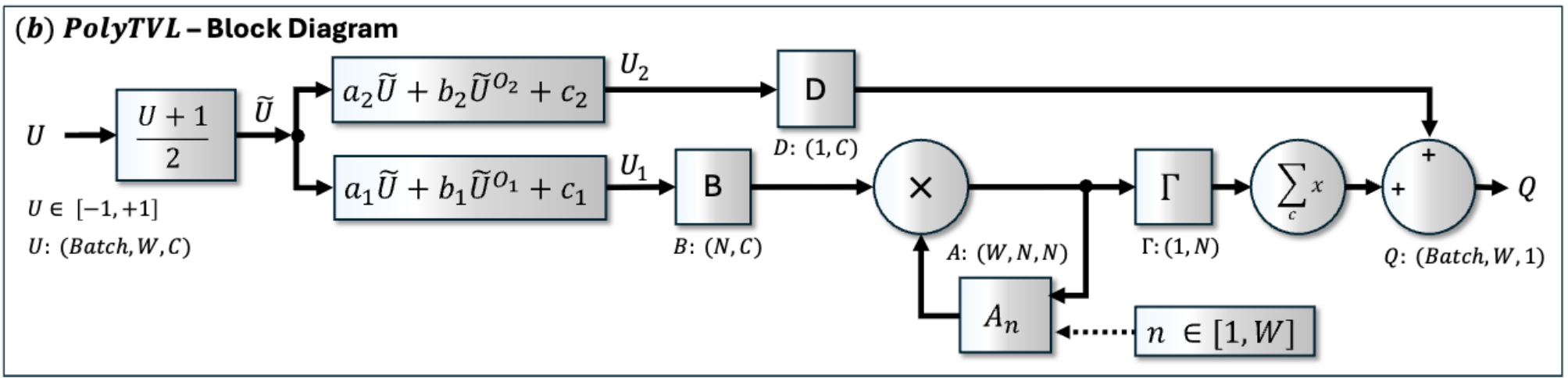}    
    \vspace{-5pt}
    \caption{(a) Proposed 2-block architecture: Feature Extraction Block (session-specific) and Decoder Block ($v01$: PolyTVL+LSTM), connected via a \textit{Hook} layer, with ablation variants ($v00$: PolyTVL+Dense, $v02$: S4D-Lin+Dense and $v03$: LSTM+Dense). (b) PolyTVL layer block-diagram.}
    \label{Fig2}
\end{figure*}    
\vspace{-5pt}
\subsection{Model Architecture}
The proposed model, implemented using TensorFlow/Keras, follows a modular two-block design consisting of a `Feature Extraction Block' and a `Decoder Block', connected via a pass-through \textit{Hook} layer of shape $(N, 9, W, 4)$ that enables independent deployment of each block. The complete architecture comprises 70,755 total weights (227 non-trainable) with a 0.270 MB memory footprint (Decoder: 98\%). The complete architecture is illustrated in Fig.~\ref{Fig2}.
\subsubsection{\textbf{Feature Extraction Block}}
The `Feature Extraction Block' accepts raw EEG of shape (N, C, W), with batch size N, 28 channels, and 256 samples (2.56~sec at 100 Hz). Each channel is z-scored using session-specific per-channel $\mu$ and $\sigma$ from reference windows (Section~\ref{sec:Preprocessing}).
\\
The normalized EEG passes through nASR~\cite{nASR2026}, a trainable channel-level Artifact Subspace Reconstruction layer that detects contaminated channels and reconstructs them from their clean neighbors (selected based on pairwise L2 distances). A weighted reconstruction layer scales reconstructed channels. Average re-referencing and 20\% spatial dropout are then applied across all 28 channels.
\\
The 28 channels are grouped into 9 ROIs defined in anatomical terms (Fig.~\ref{Fig1}). Each ROI is processed by an independent depth-wise convolution (kernel: $C_r \times 1$, $\tanh$, max-norm 1.0), collapsing $C_r$ channels into a single spatially filtered signal. The stacked ROI outputs are decomposed via RDWT using a Symlet-2 mother wavelet (MW) into five scales. The choice of MW is motivated by its established suitability for MI decoding~\cite{Sarkar2025}. A trainable zero-phase filter is applied in the wavelet domain~\cite{SarkarRDWT2026}, with sparsity regularization applied to the $\gamma$ band (25--50 Hz) via masking. Band-specific kernel sizes are assigned as: $\delta$ (99), $\theta$ (17), $\alpha$ (9), and $\beta$ (5), reflecting the inverse relationship between frequency and filter length. Inverse RDWT reconstructs four sub-bands: $\delta,\theta$ (0--6.25 Hz), $\alpha$ (6.25--12.5 Hz), $\beta$ (12.5--25.0 Hz), and $\gamma$ (25--50 Hz)-- yielding output shape (N, 9, W, 4).
\vspace{-5pt}
\subsection{\textbf{Decoder Block}}
The `Decoder Block' receives extracted features via a pass-through \textit{Hook} layer. Only three sub-bands: $\delta,\theta$ (0--6.25 Hz), $\alpha$ (6.25--12.5 Hz), and $\beta$ (12.5--25.0 Hz) are processed in the `Decoder Block' as independent parallel branches -- each branch comprising 6 stages.
\\
The first three layers follow the EEGNet architecture~\cite{Lawhern2018}: (1) a 2D Convolution (8 filters, $(9 \times 64)$ kernel) spans all 9 ROIs, followed by Batch Normalization (BN) and Dropout (20\%); (2) a Depthwise Convolution ($(9 \times 1)$ kernel, depth multiplier D=2, max-norm 1.0) captures cross-ROI interactions, followed by BN and Dropout (20\%); and (3) a Separable Convolution (16 filters, $(1 \times 64)$ kernel) with Dropout (20\%) refines features, squeezing the last axis to yield shape (N, W, 16). All three layers apply $\tanh$ activation to constrain output to $[-1,1]$, ensuring bipolar representation and numerical stability.
\\
In the 4th stage, the squeezed output is passed through the proposed PolyTVL layer with 8 hidden states. The input $U \in [-1,1]$ is shift-scaled to $\tilde{U} \in [0,1]$ for polynomial stability via:
$$
 \tilde{U} = \frac{U + 1}{2}, \quad U \in [-1, 1] \eqno{(1)}
$$
Then two learnable polynomial transforms are applied:
$$
 U_1 = a_1\tilde{U} + b_1\tilde{U}^{O_1} + c_1 \eqno{(2)}
$$
$$
 U_2 = a_2\tilde{U} + b_2\tilde{U}^{O_2} + c_2 \eqno{(3)}
$$
Here $O_1, O_2 \geq 1$ are learnable polynomial orders (initialized to~1); $a_1, a_2 \in \mathbb{R}$ and $b_1, b_2 \in \mathbb{R}$ are linear and polynomial scaling weights (initialized to~1 and~0, respectively); and $c_1, c_2 \in \mathbb{R}$ are bias terms (initialized to~0).
\\
In contrast to SSM variants~\cite{Gu2022}, which employ LTI state-space dynamics, PolyTVL introduces two key extensions: (i) dual learnable polynomial transforms with positive-constrained orders $O_1, O_2 \geq 1$ for nonlinear dynamics, and (ii) a time-varying state transition matrix $A \in \mathbb{R}^{(W \times N \times N)}$, enabling position-specific state dynamics at each timestep. The PolyTVL output $Q_n$ at a particular timestep is computed as:
$$
 Q_n = \sum_{c=1}^{C}(\Gamma \cdot A_n \cdot B)_c \cdot U_{1,(n,c)} 
        + \sum_{c=1}^{C} D_c \cdot U_{2,(n,c)} \eqno{(4)}
$$
\noindent where:\\[4pt]
\begin{tabular}{ll}
$A \in \mathbb{R}^{W \times N \times N}$  & : time-varying state transition matrix,\\
$A_n \in \mathbb{R}^{N \times N}$         & : $n$-th slice of $A$,\\
$B \in \mathbb{R}^{N \times C}$         & : input projection matrix,\\
$\Gamma \in \mathbb{R}^{1 \times N}$    & : output projection matrix,\\
$D \in \mathbb{R}^{1 \times C}$         & : skip connection matrix,\\
$N = 8$                                  & : number of hidden states,\\
$W$                                      & : input sequence length,\\
$C$                                      & : number of input channels.
\end{tabular}
\\[4pt]
\\
In $5^{th}$~stage, PolyTVL output is down-sampled via a 1D Average Pooling (pool size=2 and stride=2), reducing the temporal dimension to $(N, W/2, 1)$, followed by an LSTM layer ($6^{th}$~stage) with 8 units that compresses the sequence into a compact fixed-length vector $(N, 8)$. The band-specific vectors are then concatenated and passed through a Dense layer (16 units, ReLU), followed by a 4-unit Dense layer with a softmax activation predicting four gait states: \textit{Stand}, \textit{Initiate}, \textit{Execute}, and \textit{Terminate}.
\\
Four variants of the decoder (Fig.~\ref{Fig2}) were evaluated: $v00$ (PolyTVL+Dense), $v01$ (PolyTVL+LSTM, proposed), $v02$ (S4D-Lin+Dense) and $v03$ (LSTM+Dense). All variants use 8 latent states (units); in $v00$, $v02$ and $v03$, a Dense(1) layer summarizes sub-bands into a single feature, while all other components remained identical.
\vspace{-10pt}
\begin{table}[h]
\centering
\caption{Confusion-Aware Penalty Matrix}
\vspace{1pt}
\label{tab:Table_1}
\renewcommand{\arraystretch}{0.8}
\setlength{\tabcolsep}{4pt}
{\small
\begin{tabular}{lcccc}
\hline
\noalign{\vspace{2pt}}
& \textbf{Terminate} & \textbf{Stand} & \textbf{Execute} & \textbf{Initiate} \\
\hline
\noalign{\vspace{2pt}}
\textbf{Terminate} & 0  & 5  & 20 & 30 \\
\textbf{Stand}     & 5  & 0  & 15 & 20 \\
\textbf{Execute}   & 20 & 15 & 0  & 10 \\
\textbf{Initiate}  & 30 & 20 & 10 & 0  \\
\hline
\end{tabular}
}
\vspace{-5pt}
\end{table}
\begin{figure*}[!t]
    \centering
    \includegraphics[width=0.9\textwidth]{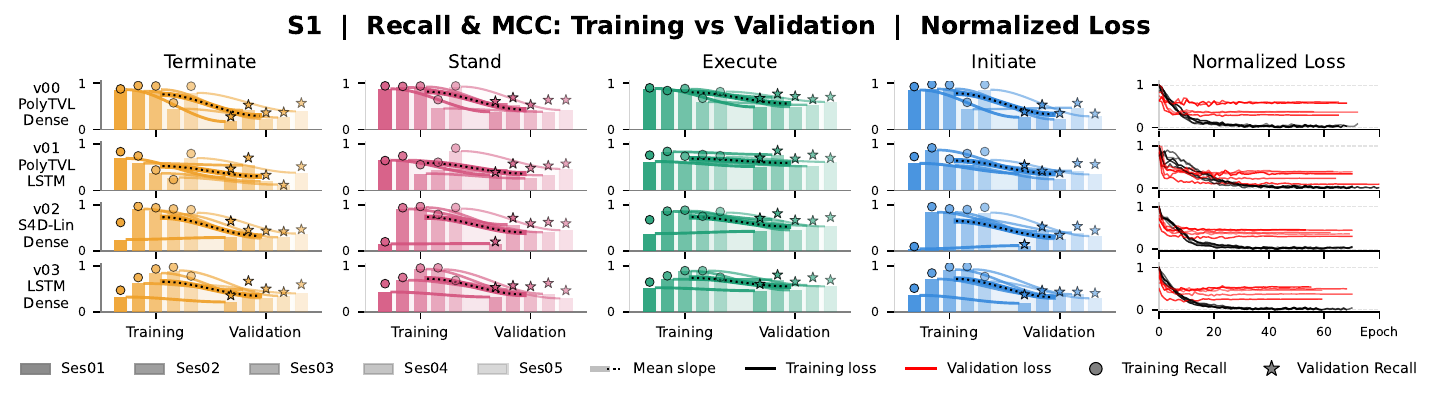}
    \vspace{-10pt}
    \caption{Normalized Recall, MCC, and Loss (Training vs. Validation) across Ses.~1--5 for ablation: $v00$, $v01$, $v02$ and $v03$.}
    \label{Fig3}
\end{figure*}
\begin{table*}[h]
\centering
\vspace{-8pt}
\caption{Learned PolyTVL polynomial transformation weights ($\mu \pm \sigma$, OL Ses.~1--5) per frequency band.}
\vspace{2pt}
\label{tab:Table_2}
\small
\renewcommand{\arraystretch}{1.0}
\setlength{\tabcolsep}{4pt}
\begin{tabular}{lcccccccc}
\hline
\textbf{Band} & $O_1$ & $a_1$ & $b_1$ & $c_1$ & $O_2$ & $a_2$ & $b_2$ & $c_2$ \\
\hline
$\delta,\theta$ & 1.068 (0.070) & 1.138 (0.036) & 0.139 (0.037) & -0.079 (0.048) & 1.041 (0.045) & 1.138 (0.021) & 0.139 (0.021) & -0.044 (0.026) \\
$\alpha$        & 1.115 (0.022) & 1.170 (0.023) & 0.174 (0.025) & -0.018 (0.015) & 1.064 (0.017) & 1.108 (0.014) & 0.109 (0.014) &  0.015 (0.010) \\
$\beta$         & 1.205 (0.106) & 1.213 (0.037) & 0.217 (0.039) & -0.106 (0.069) & 1.084 (0.078) & 1.019 (0.088) & 0.023 (0.089) & -0.119 (0.061) \\
\hline
\end{tabular}
\end{table*}
\subsection{Training Strategy}
Training used the Adam optimizer (lr = $10^{-3}$, batch size = 128) for up to 200 epochs, with early stopping (patience = 50) and lr halving on plateau (patience = 5, min = $10^{-7}$). The loss combined categorical cross-entropy with a penalty $P[\hat{y}, y]$ derived from a custom $4\times4$ matrix (Table~\ref{tab:Table_1}) reflecting the gait sequence: \textit{Stand} $\rightarrow$ \textit{Initiate} $\rightarrow$ \textit{Execute} $\rightarrow$  \textit{Terminate}. Adjacent confusions were lightly penalized, whereas non-adjacent confusions received the most severe penalties. Maximum penalties were assigned to \textit{Initiate}/\textit{Terminate} misclassifications to de-risk erroneous transitions. Hyperparameters were set heuristically; systematic optimization is future scope.
\\
The complete model was trained end-to-end for each OL Ses.~1-5 independently. Features from all five session-specific models were then aggregated to retrain the `Decoder Block', whose weights were subsequently frozen. From Ses.~6 onward, only the `Feature Extraction Block' was retrained at each session using OL Runs~1-3 for session-specific adaptation. A consistent 75:25 train/validation split was used throughout, with steps 1-10 and 16-20 for training and steps 11-15 for validation. Performance was evaluated during CL Runs~4-9 using the session-specific `Feature Extraction Block' paired with the frozen `Decoder Block'.
\begin{table}[!h]
\centering
\vspace{-5pt}
\caption{CL Gait Initiation Success Rate (\%)}
\vspace{2pt}
\label{tab:Table_3}
\renewcommand{\arraystretch}{0.8}
\setlength{\tabcolsep}{3pt}
{\small
\begin{tabular}{l|ccc|ccc}
\hline
\noalign{\vspace{2pt}}
\multirow{2}{*}{Ses.} & \multicolumn{3}{c|}{CL Rex} & \multicolumn{3}{c}{CL Walk} \\
 & Run-4 & Run-5 & Run-6 & Run-7 & Run-8 & Run-9 \\
\hline
\noalign{\vspace{2pt}}
  6 & 70 & 60 & 90 & 50 & 50 & 20 \\
  7 & 90 & 30 & 50 & 50 & 20 & 80 \\
  8 & 90 & 40 & 70 & 70 & 60 & 40 \\
  9 & 60 & 60 & 20 & 60 & 80 & 80 \\
  10 & 40 & 20 & 40 & 10 & 70 & 50 \\
\hline
\end{tabular}
}
\vspace{-5pt}
\end{table}
\begin{figure}[!ht]
    \centering
    \includegraphics[width=\linewidth]{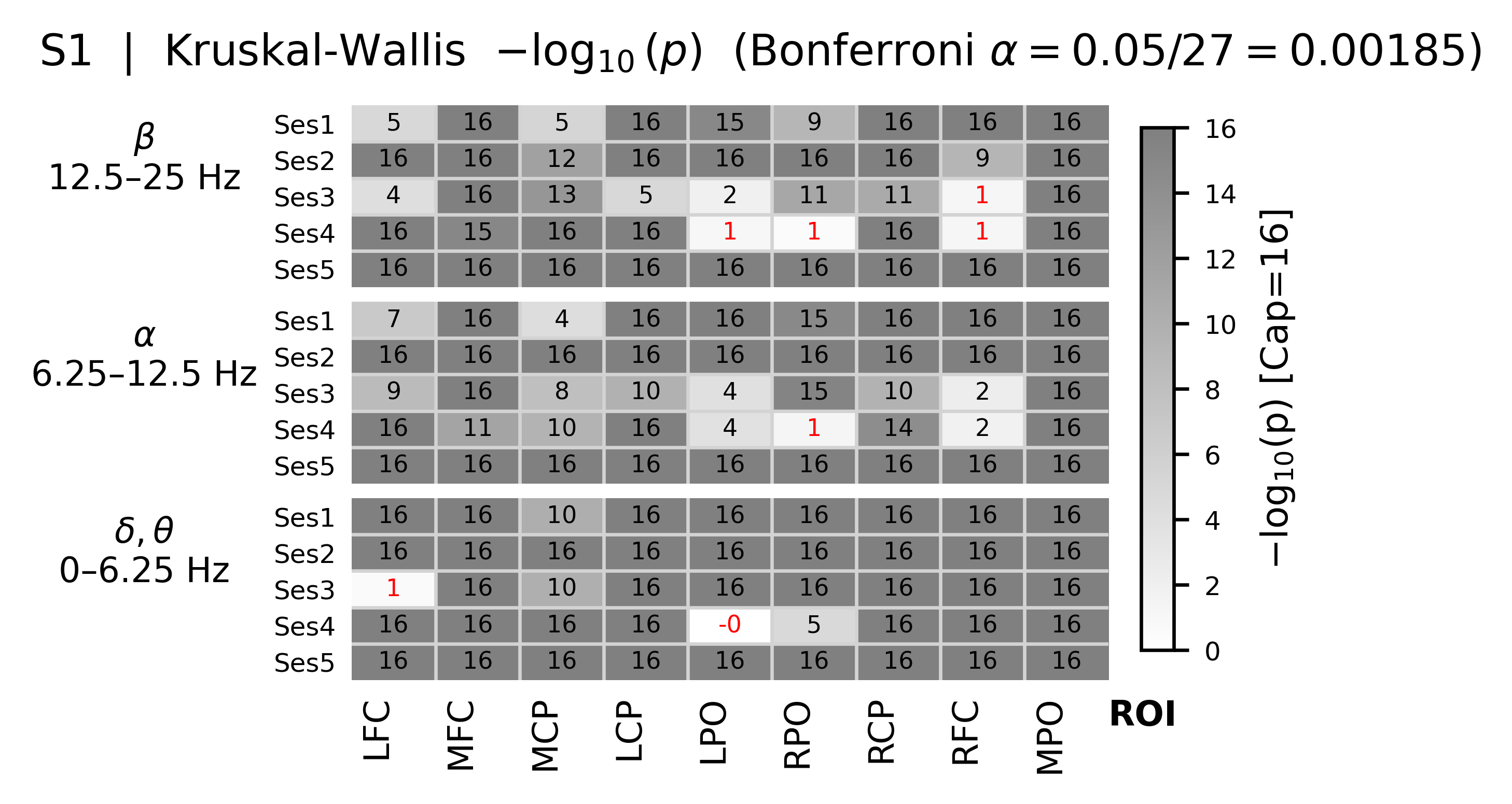}
    \vspace{-20pt}
    \caption{Kruskal-Wallis $-\log_{10}(p)$ (Ses.~1--5), Bonferroni corrected ($\alpha=0.05/27$; 9~ROIs~$\times$~3~Bands). \textcolor{red}{Red}: $p>0.05$.}
    \label{Fig4}
\vspace{-15pt}    
\end{figure}
\vspace{-10pt}
\section{Results}
The mean 95\% CI Recall chance thresholds were \textit{Stand} 25.2\%, \textit{Initiate} 20.0\%, \textit{Execute} 50.5\%, and \textit{Terminate} 20.0\%. Matthews Correlation Coefficient (MCC), which accounts for all confusion matrix terms, was used for overall evaluation. Fig.~\ref{Fig3} shows Recall and MCC across Ses.~1--5 alongside training/validation loss for all 
four variants.
\\
$v00$ had the largest overfitting gap (MCC gap: 0.362) and 
$v01$ the lowest (0.187). $v01$ achieved the highest validation 
MCC (0.435) and strongest \textit{Execute} (0.594) and 
\textit{Initiate} (0.359) scores across all variants. Ses.~3 underperformed, as the participant rushed to finish early. $v01$ was selected for CL.
\\
Kruskal-Wallis analysis (Fig.~\ref{Fig4}) revealed significant discriminability ($p<0.05$, Bonferroni corrected, $\alpha=0.05/27$) across gait classes in 128/135 ROI-session-band combinations. $\delta,\theta$ yielded the highest $-\log_{10}(p)$ overall; MFC, MCP, LCP, RCP, and MPO remained discriminative across all bands and sessions. Table~\ref{tab:Table_2} summarizes learned polynomial transformation weights of PolyTVL ($\mu~\pm~\sigma$, OL Ses.~1-5). $O_1$ increases monotonically ($\delta,\theta$: 1.068, $\alpha$: 1.115, $\beta$: 1.205), confirming progressively stronger nonlinear reshaping at higher frequencies.
\\
Table~\ref{tab:Table_3} reports gait initiation success rates ($\geq$2 correct predictions in the \textit{Initiate} window) for Ses.~6--10. Success averaged 55.3\% for REX-assisted Runs~4--6 and 52.7\% for volitional Runs~7--9. For REX-assisted runs, per-window TPR and FPR were 26.8\% and 16.7\%, respectively, yielding above-chance performance (Youden's $J=10.1\%$, 95\% CI [7.6, 12.5], $z=9.14$, $p<0.001$). Ses.~10 showed the lowest performance, coinciding with participant haste.
The mean end-to-end processing time, including preprocessing and prediction, was 70.5~ms ($\pm$41.5) on an Intel i7-14650HX (32~GB RAM), thereby validating real-time feasibility. For valid REX-assisted initiations (n = 83), command and REX latencies were 784~$\pm$~401 ms and 868~$\pm$~405 ms, respectively.
\vspace{-5pt}
\section{Conclusion}
We proposed a 2-block BCI architecture -- a trainable `Feature Extraction Block' enabling real-time denoising and multi-domain feature extraction along with a PolyTVL+LSTM `Decoder Block' for four-state EEG-based gait classification. Ablation confirmed $v01$ (validation MCC: 0.435, gap: 0.187) with consistent feature discriminability across ROIs and sub-bands ($p < 0.05$). CL deployment on a Rex exoskeleton achieved 55.3\% gait initiation success, validating feasibility in this single-subject pilot. Future work will extend to multi-subject cohorts.

\begingroup
\small
\setlength{\parskip}{0pt}

\endgroup


\begin{thebibliography}{99}

\bibitem{Contreras2016}
J. L. Contreras-Vidal et al.,
``Powered exoskeletons for bipedal locomotion after spinal cord injury,''
\emph{Journal of Neural Engineering},
vol. 13, no. 3, Art. no. 031001, 2016.

\bibitem{Pfurtscheller1994}
G. Pfurtscheller and C. Neuper,
``Event-related synchronization of mu rhythm in the EEG over the cortical
hand area in man,''
\emph{Neuroscience Letters},
vol. 174, no. 1, pp. 93--96, 1994.

\bibitem{Ortiz2023}
M. Ortiz, K. Nathan, J. M. Azor\'in, and J. L. Contreras-Vidal,
``Brain-machine interfaces for neurorobotics,''
in \emph{Handbook of Neuroengineering},
N. V. Thakor, Ed.
Singapore: Springer, 2023, pp. 1817--1857.

\bibitem{MorenoCastelblanco2025}
S. R. Moreno-Castelblanco, M. A. V\'elez-Guerrero, and
M. Callejas-Cuervo,
``Artificial intelligence approaches for EEG signal acquisition and
processing in lower-limb motor imagery: A systematic review,''
\emph{Sensors},
vol. 25, Art. no. 5030, 2025.

\bibitem{McDermott2022}
E. J. McDermott, P. Raggam, S. Kirsch, P. Belardinelli, U. Ziemann,
and C. Zrenner,
``Artifacts in EEG-based BCI therapies: Friend or foe?''
\emph{Sensors},
vol. 22, no. 1, Art. no. 96, 2022.

\bibitem{Roeder2024}
L. Roeder, M. Breakspear, G. K. Kerr, and T. W. Boonstra,
``Dynamics of brain-muscle networks reveal effects of age and
somatosensory function on gait,''
\emph{iScience},
vol. 27, no. 3, Art. no. 109162, Mar. 2024.

\bibitem{He2014}
Y. He et al.,
``An integrated neuro-robotic interface for stroke rehabilitation using
the NASA X1 powered lower limb exoskeleton,''
in \emph{Proc. IEEE Engineering in Medicine and Biology Society (EMBC)},
2014, pp. 3985--3988.

\bibitem{Nakagome2020}
S. Nakagome, T. P. Luu, Y. He, A. S. Ravindran, and
J. L. Contreras-Vidal,
``An empirical comparison of neural networks and machine learning
algorithms for EEG gait decoding,''
\emph{Scientific Reports},
vol. 10, no. 1, Art. no. 4372, 2020.

\bibitem{Tortora2020}
S. Tortora, S. Ghidoni, C. Chisari, S. Micera, and F. Artoni,
``Deep learning-based BCI for gait decoding from EEG with LSTM recurrent
neural network,''
\emph{Journal of Neural Engineering},
vol. 17, no. 4, Art. no. 046011, 2020.

\bibitem{Gu2022}
A. Gu, K. Goel, A. Gupta, and C. R\'e,
``On the parameterization and initialization of diagonal state space
models,''
in \emph{Advances in Neural Information Processing Systems (NeurIPS)},
vol. 35, 2022, pp. 35971--35983.

\bibitem{Guo2025}
M. Guo et al.,
``MI-Mamba: A hybrid motor imagery electroencephalograph classification
model with Mamba's global scanning,''
\emph{Annals of the New York Academy of Sciences},
vol. 1544, no. 1, pp. 242--253, 2025.

\bibitem{Rashmi2022}
C. R. Rashmi and C. P. Shantala,
``EEG artifacts detection and removal techniques for brain--computer
interface applications: A systematic review,''
\emph{International Journal of Advanced Trends in Engineering and
Technology}, 2022.

\bibitem{nASR2026}
S. Sarkar and J. L. Contreras-Vidal,
``nASR: An end-to-end trainable neural layer for channel-level EEG
artifact subspace reconstruction in real-time BCI,''
\emph{arXiv preprint arXiv:2605.14941}, 2026.

\bibitem{HInf2016}
A. Kilicarslan, R. G. Grossman, and J. L. Contreras-Vidal,
``A robust adaptive denoising framework for real-time artifact removal
in scalp EEG measurements,''
\emph{Journal of Neural Engineering},
vol. 13, no. 2, Art. no. 026013, 2016.

\bibitem{Singh2021}
A. Singh, A. A. Hussain, S. Lal, and H. W. Guesgen,
``A comprehensive review on critical issues and possible solutions of
motor imagery based electroencephalography brain-computer interface,''
\emph{Sensors},
vol. 21, no. 6, Art. no. 2173, 2021.

\bibitem{Zhang2017}
Y. Zhang et al.,
``Multiple kernel based region importance learning for neural
classification of gait states from EEG signals,''
\emph{Frontiers in Neuroscience},
vol. 11, Art. no. 170, 2017.

\bibitem{Pooja2022}
Pooja, S. Pahuja, and K. Veer,
``Recent approaches on classification and feature extraction of EEG
signal: A review,''
\emph{Robotica},
vol. 40, no. 1, pp. 77--101, 2022.

\bibitem{SarkarRDWT2026}
S. Sarkar, S. S. Gandavarapu, J. Feng, S. Prasad, R. Khanbabaie,
and J. L. Contreras-Vidal,
``BCI-based assessment of ocular response time using dynamic time
warping leveraging an RDWT-driven deep neural framework,''
\emph{arXiv preprint arXiv:2605.14883}, 2026.

\bibitem{Pope2022}
K. J. Pope et al.,
``Managing electromyogram contamination in scalp recordings: An
approach identifying reliable beta and gamma EEG features of psychoses
or other disorders,''
\emph{Brain and Behavior},
vol. 12, no. 9, Art. no. e2721, 2022.

\bibitem{shibasaki2006}
H. Shibasaki and M. Hallett,
``What is the Bereitschaftspotential?''
\emph{Clinical Neurophysiology},
vol. 117, no. 11, pp. 2341--2356, 2006.

\bibitem{Bohannon2015}
R. W. Bohannon, Y. C. Wang, and R. C. Gershon,
``Two-minute walk test performance by adults 18 to 85 years: Normative
values, reliability, and responsiveness,''
\emph{Archives of Physical Medicine and Rehabilitation},
vol. 96, no. 3, 2015.

\bibitem{Sarkar2025}
S. Sarkar and J. L. Contreras-Vidal,
``Optimal discrete mother wavelet selection for EEG motor imagery
decoding: A comparative study,''
in \emph{Health Informatics and Medical Systems and Biomedical
Engineering}, ser. Communications in Computer and Information Science,
vol. 2259.
Springer, 2025.

\bibitem{Lawhern2018}
V. J. Lawhern, A. J. Solon, N. R. Waytowich, S. M. Gordon,
C. P. Hung, and B. J. Lance,
``EEGNet: A compact convolutional neural network for EEG-based
brain--computer interfaces,''
\emph{Journal of Neural Engineering},
vol. 15, no. 5, Art. no. 056013, 2018.

\end{thebibliography}
\end{document}